\documentclass{article}

\usepackage[preprint]{corl_2025} 

\usepackage{amsmath} 
\usepackage{amssymb}  
\usepackage{dsfont}
\usepackage{graphicx}
\usepackage{float}
\usepackage{placeins}
\usepackage{multirow}
\usepackage{pgf,tikz}
\usepackage{nowidow}
\usepackage{lineno}
\usepackage{tabularx}
\usepackage{booktabs}
\usepackage{caption}
\usepackage{subcaption}

\DeclareMathOperator*{\argmin}{arg\,min}

\usepackage{siunitx}
\usepackage{color}
\usepackage{flushend}
\usepackage{lineno}
\usepackage{algorithm}
\usepackage[noend]{algorithmic}
\allowdisplaybreaks

\newcommand{\update}[1]{\textcolor{black}{#1}}

\title{In-Context Iterative Policy Improvement for Dynamic Manipulation}

\author{
  Mark Van der Merwe\\
  Department of Robotics, University of Michigan\\
  \texttt{markvdm@umich.edu} \\
  \And
  Devesh K. Jha \\
  MERL\\
  \texttt{jha@merl.com} \\
}

\begin{document}

\maketitle


\begin{abstract}
Attention-based architectures trained on internet-scale language data have demonstrated state of the art reasoning ability for various language-based tasks, such as logic problems and textual reasoning. Additionally, these Large Language Models (LLMs) have exhibited the ability to perform few-shot prediction via in-context learning, in which input-output examples provided in the prompt are generalized to new inputs. This ability furthermore extends beyond standard language tasks, enabling few-shot learning for general patterns. In this work, we consider the application of in-context learning with pre-trained language models for dynamic manipulation. Dynamic manipulation introduces several crucial challenges, including increased dimensionality, complex dynamics, and partial observability. To address this, we take an iterative approach, and formulate our in-context learning problem to predict adjustments to a parametric policy based on previous interactions. We show across several tasks in simulation and on a physical robot that utilizing in-context learning outperforms alternative methods in the low data regime. Video summary of this work and experiments can be found \href{https://youtu.be/2inxpdrq74U?si=dAdDYsUEr25nZvRn}{here}.
\end{abstract}
\section{Introduction}\label{sec:intro}

Transformer-based architectures trained on internet-scale data, Large Language Models (LLMs), have recently become a powerful tool for a variety of language and image-based tasks~\cite{yang_dawn_2023}. Recent works have proposed robotics methods that exploits the reasoning ability of these LLMs over textual and visual inputs to perform high-level planning~\cite{ichter_saycan_2023}, generate robot control code~\cite{liang_cap_2023}, and derive reward functions~\cite{huang_voxposer_2023}. These methods all exploit \textit{in-weights learning}, exploiting the information stored in the weights of the networks. But what if a task is not well reflected in textual inputs or even in visual information?

Amongst the emergent abilities, these Large Language Models (LLMs) can perform few-shot learning, in which a small number of input-output examples are provided in the prompt to the model, before being provided with a test input~\cite{brown_LLMFSL_2020}. This form of learning has been dubbed \textit{in-context learning}. Curiously, this ability has been shown to extend to patterns that are not necessarily reflected in the training domain, suggesting that LLMs act as \textit{general pattern machines}~\cite{mirchandani_GPM_2023}. This invites an intriguing proposition: that large-scale language pre-training can enable few-shot learning in new domains. In this work, we explore how in-context learning can be applied for a class of robotic tasks which are not easy to reflect in textual and visual inputs alone: dynamic manipulation.

Robotic manipulation that exploits \textit{dynamics} can introduce several benefits over quasi-static systems. Dynamic manipulation can increase the robot workspace~\cite{zhang_arc_2021, zeng_tossingbot_2020}, improve task efficiency~\cite{ha_flingbot_2022,xu2022dextairity}, and extend robot capabilities with new skills such as dynamic in-hand re-grasping~\cite{wang_swingbot_2020}. Designing systems for dynamic manipulation introduces several challenges. First, the system dynamics are often dependent upon underlying physical properties that are not easy to observe directly, including from vision, such as the mass of an object, friction of a surface, or density of a rope. These properties require dynamic interactions~\cite{xu_densephysnet_2019,wang2025implicit} and/or additional sensing~\cite{wang_swingbot_2020} to accurately observe. Second, the increased dimensionality of the system (such as deformable manipulation~\cite{ha_flingbot_2022, chi_irp_2022}) and the complexity of the dynamics (such as frictional contact interactions) have largely limited current systems to single-task solutions. These solutions generally involve collecting large datasets and training task-specific solutions~\cite{chi_irp_2022,xu_densephysnet_2019,lim2022real2sim2real}.

\begin{figure}
    \centering
    \includegraphics[width=\linewidth]{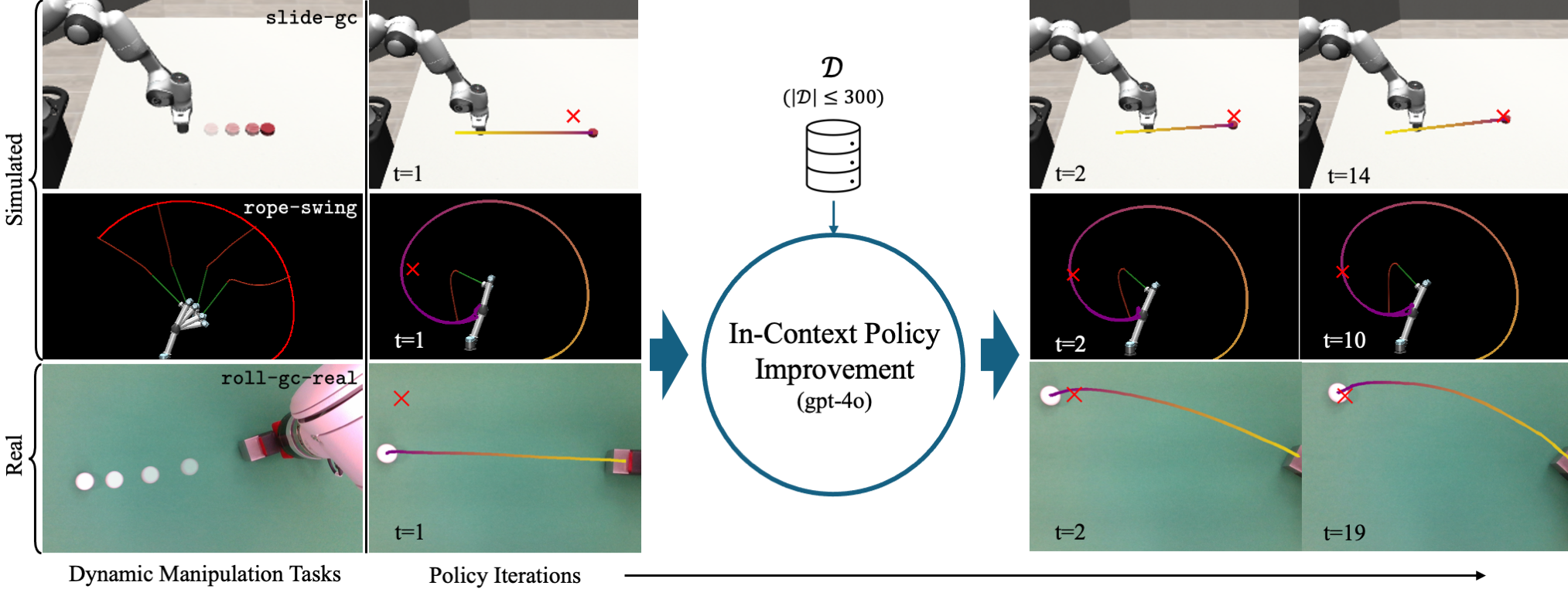}
    \caption{We investigate in-context learning for iteratively improving policy parameters for dynamic manipulation tasks. We demonstrate across a variety of tasks, both in simulation and on a real robot, that utilizing \textit{in-context learning} with a pretrained Large Language Model (such as OpenAI's gpt-4o), can learn to improve dynamic manipulation policies interactively from a small ($\leq 300$) policy improvement dataset.}
    \label{fig:teaser}
\end{figure}

Humans performing dynamic manipulation may fail on a first attempt, but utilize their experience interacting with similar dynamical systems to iteratively improve their approach on a target system. In this work, we seek to enable robots that also can iteratively improve their interactions with a dynamical system. We investigate enabling a few-shot iterative policy improvement technique, utilizing in-context-learning with pre-trained LLMs. We formulate our problem as learning an improvement operator, which predicts adjustments to a parametric policy based on the results of previous interactions. Previous interactions are tokenized and fed as inputs along with policy parameters and in-context learning is utilized to output a change to the policy parameters that drives towards task success. We utilize a small dataset of improvement labels from interactions with similar tasks to form our in-context examples. 

We apply our method to a variety of dynamic manipulation tasks, in simulation and on a physical robot (Fig.~\ref{fig:teaser}). \update{Our results indicate the novel ability of pre-trained LLMs to perform in-context policy improvement for dynamic manipulation tasks without requirement of any fine-tuning or training.} We show that our proposed approach outperforms utilizing in-weights reasoning or alternative in-context policy approaches~\cite{mirchandani_GPM_2023} and compares favorably to alternative policy optimization approaches (e.g., Bayesian Optimization) and \update{other policy improvement operators in the low data regime.} 

\section{Related Work}\label{sec:related_works}

\subsection{Large Language Models and In-Context Learning}

Large Language Models (LLMs), attention-based architectures~\cite{NIPS2017_3f5ee243} trained on internet-scale language data, have demonstrated remarkable ability in generating solutions to various language-based tasks such as logic problems and math puzzles~\cite{srivastava2023beyond,suzgun2022challenging} and reasoning about joint visual and textual inputs~\cite{yang_dawn_2023}. As model scale increased, LLMs were shown to be capable of few-shot learning, where input-output examples for an unseen problem are provided in the prompt, without the need for model updates~\cite{brown_LLMFSL_2020}. LLMs thus exhibit two forms of learning: \textit{in-weights learning} involves information stored in the model weights via gradient updates on the training dataset, while \textit{in-context learning} involves information provided only in the inputs at inference time (i.e., the prompt), with no weight updates. Recent work has shown that \textit{in-context learning} can be applied outside of the training domain, suggesting that pre-trained LLMs can act as general pattern machines, capable of identifying input-output patterns in-context in new target domains~\cite{mirchandani_GPM_2023}.

In-context learning (ICL) emerges as a result of explicitly training on large contexts that cover multiple task examples~\cite{brown_LLMFSL_2020} or implicitly due to distributional properties of the training data~\cite{NEURIPS2022_77c6ccac}. Some work suggests ICL is driven by induction heads, an attention mechanism which associates completions with previous, related completions in the prompt~\cite{olsson2022context}. Behaviorally, transformers exhibit examplar-based generalization in-context, compared to rule-based generalization in-weights~\cite{chan2022transformers}. The phenomena is still not fully understood: \citet{min2022rethinking} showed that for certain text-based tasks, randomizing the output labels in the context has little effect on performance, suggesting that exposure to input/output distributions and/or improved in-weights recall drives improvement, rather than learning from in-context labeled data. ICL has, however, been demonstrated on out of distribution tasks, suggesting labels are utilized in certain cases~\cite{NEURIPS2022_77c6ccac,mirchandani_GPM_2023}.

\subsection{LLM for Robotics}

The majority of LLM-enabled robotics methods to date largely exploit in-weights reasoning. These systems utilize the reasoning capabilities of large pre-trained models to perform zero-shot (or few-shot) reasoning on a variety of robotics tasks by prompting LLMs to generate high-level plans~\cite{ichter_saycan_2023, sun2024interactive}, robot control code~\cite{liang_cap_2023}, value functions for trajectory optimization~\cite{huang_voxposer_2023,yu_lang2reward_2023}, reward functions for Reinforcement Learning (RL)~\cite{wang_rlvlmf_2024}, and motion trajectories~\cite{kwon_ztg_2024}. These methods demonstrate flexible systems that can be applied to broad classes of tasks, purely exploiting the knowledge and reasoning in the pretrained models. The tasks considered, however, are largely quasi-static and kinematic in nature, such as pick-and-place, opening/closing, and object retrieval. The emphasis on largely high-level reasoning tasks may result in more subject matter overlap in an LLM training corpus. Reasoning about low-level dynamics, however, is unlikely to appear in the training data as the relevant data is rarely presented explicitly, let alone in text. This motivates our investigation of ICL as an LLM mechanism applicable to dynamic manipulation.

\subsection{In-Context Learning for Robotics}

Most related to our work is the sequence improvement demonstrated in~\cite{mirchandani_GPM_2023}. They propose a purely in-context method that improves a trajectory for tasks such as cartpole or a reaching task, where the trajectory/policy is tokenized and iteratively improved based on execution feedback. In contrast, our method uses a small number of examples to fit a policy improvement operator in context. Also related is Keypoint Action Tokens, which performs behavior cloning via in-context learning~\cite{di_keypoint_2024}. Behavior cloning is difficult to apply to these dynamic manipulation tasks as the task parameters are not always available a priori. \update{Finally, some work trains transformers from scratch for multi-task policy adaptation~\cite{yoo2025context} or system identification~\cite{zhang2025dynamics}. This requires significant data and modeling effort. In contrast, our work relies solely on learning in-context using a pre-trained transformer.}

\section{Problem Formulation}\label{sec:problem_statement}

We consider solving a dynamic manipulation task drawn from a task distribution $\tau \sim P(\tau)$. $\tau$ reflects the variation in the task, such as physical parameters of the system (e.g., friction or mass) or the desired goal for a goal-conditioned policy. We assume a parametric policy $\pi_{\theta}$. We have access to the environment through $s_{1:T}=T_\tau(\pi_{\theta}, s_0)$, where $s_t$ is the state of the system at time $t$ and $s_0$ is the initial state. Finally, we have access to a task cost function $C_{\tau}(s_{1:T})$ which evaluates the cost of a particular rollout. The objective is to identify the best policy parameters for the given task:
\begin{equation} \label{eq:policy_min}
    \theta^* = \argmin_{\theta} C_\tau(T_\tau(\pi_{\theta}, s_0))
\end{equation}

\subsection{Tasks}
We consider five different tasks using three robot setups, two simulated and one physical robot. We describe in each case the task parameters, state definition, and policy parameterization. In accordance with similar dynamic manipulation work, we utilize parametric motions as our policy, designed in accordance with each task~\cite{chi_irp_2022}. The task setups are shown in Fig.~\ref{fig:teaser}.

\subsubsection{Slide Task (\texttt{slide})} In this simulated task, a cylindrical puck is placed on a support surface in front of a Franka Emika Panda robot manipulator with a cylindrical end effector attachment. The task is to strike the puck with the robot end effector, causing it to slide along the surface to some goal configuration. In this task iteration, the start and goal location is fixed, while the physical parameters of the puck are adjusted. In particular, we adjust the puck radius $\tau_r$ and the surface friction coefficient $\tau_\mu$. The task parameters $\tau_r, \tau_\mu$ are unknown to the policy. Our state space $s_t\in \mathbb{R}^2$ is the location of the puck center at time $t$ on the plane of the table. The goal $\tau_g\in \mathbb{R}^2$ is a fixed location on the table. The task cost function is the distance to the goal configuration,
\begin{equation}
    C^{\texttt{slide}}_\tau (s_{1:T}) = ||s_T - \tau_g||_2
\end{equation}
For this task, we use a 3-dimensional policy parameterization $\theta=(\theta_{\alpha}, \theta_{d},\theta_{t})$ which defines the robot motion. $\theta_{\alpha}$ is the angle of the robot motion from start, with 0 degrees pushing straight forward. $\theta_d$ is the distance to move during the push. $\theta_t$ is the time to complete the push.

\subsubsection{Slide Goal-Conditioned Task (\texttt{slide-gc})} This is a goal-conditioned variation on the simulation slide task. We fix the puck radius and friction and instead adjust the goal configuration $\tau_g \in \mathbb{R}^2$. We use the same task cost function as the $\texttt{slide}$ task and change the policy parameterization slightly so that $\theta_{\alpha}$ is the approach angle to the puck to ensure contact is made.

\subsubsection{Rope Swing Task (\texttt{rope-swing})}

Our next task is the simulated rope swing task from Chi et al~\cite{chi_irp_2022}. In the task, a rod is attached to the end of a UR5e robot manipulator, and a rope is attached to the end of the rod. The goal of the task is to swing the rope so that the tip of the rope passes through a particular location. In this task iteration, the goal location is fixed. We adjust the rod length $\tau_r$ and rope length $\tau_l$, and both values are unknown to the policy. The state space $s_t\in \mathbb{R}^2$ is the location of the rope end in the Y-Z plane, in which the swing occurs. The goal $\tau_g\in \mathbb{R}^2$ is a fixed location in the plane. The task cost function is the smallest distance from the state to the goal during the swing,
\begin{equation}
    C_{\tau}^{\texttt{rope-swing}}(s_{1:T})=\min_{t} ||s_t - \tau_g||_2
\end{equation}
We use the same policy parameterization from Chi et al~\cite{chi_irp_2022}. Namely, we use a 3-dimensional policy parameterization $\theta=(\theta_v,\theta_{J_2}, \theta_{J_3})$, which defines the swing. $\theta_v$ is the angular velocity while $\theta_{J_2}$ and $\theta_{J_3}$ defines the end motion of the second and third joints of the robot.

\subsubsection{Rope Swing Goal-Conditioned Task (\texttt{rope-swing-gc})}

This is a goal-conditioned variation on the simulated rope swing task. We fix the rod and rope lengths and adjust the goal configuration $\tau_g\in \mathbb{R}^2$. We use the same policy and task cost parameterization as the \texttt{rope-swing} task.

\subsubsection{Real Roll Goal-Conditioned Task (\texttt{roll-gc-real})}

In this task we perform a goal-conditioned ball rolling task, using a real MELFA-Assista robot with a block end-effector for striking the ball. We place a billiards table in front of the robot and place the cue ball in a fixed position in front of the robot. The state space $s_t\in \mathbb{R}^2$ is the \textit{pixel} location of the ball tracked in a top-down camera view. The task is to strike the ball causing it to roll to the sampled goal pixel location $\tau_g\in \mathbb{R}^2$. We use the same task cost function and policy parameterization as the slide goal conditioned task (\texttt{slide-gc}).
\section{Method} \label{sec:method}

When performing a dynamic manipulation task, it is often not feasible to expect success on the first try, as important task parameters may not be readily observable a priori or due to the complexity of the task. Instead, several interactions with the system allow one to observe an outcome and correct course. For example, \citet{xu_densephysnet_2019} uses exploratory interactions to understand physical parameters before executing a dynamic manipulation task. In \citet{chi_irp_2022}, the authors propose executing a task, then learning how to adjust the trajectory to improve performance. While these demonstrate the value of iterative reasoning for dynamic tasks, they require large training datasets and complex neural architectures. We investigate if a similar iterative policy approach can be achieved utilizing the \textit{in-context learning} ability of LLMs pretrained on language data.

\subsection{Policy Improvement Operator}

\begin{figure}
    \centering
    \includegraphics[width=0.8\linewidth]{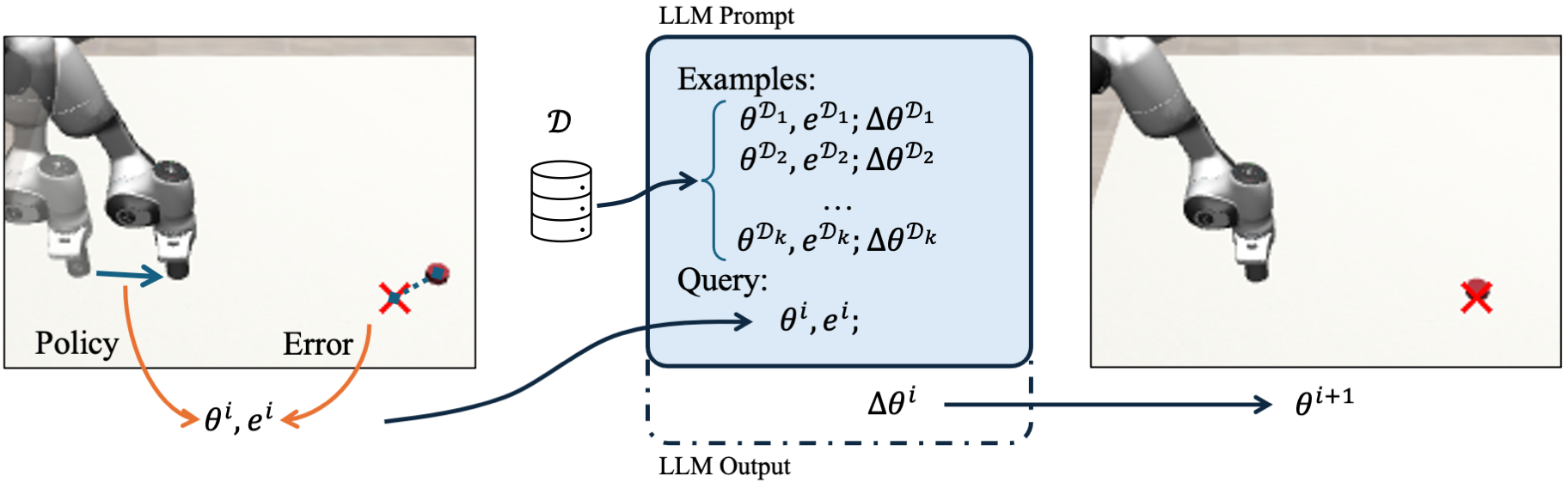}
    \caption{Overview of our proposed In-Context Policy Improvement (ICPI) method. We tokenize policy parameters and error of the current policy and provide it, along with examples from a small policy improvement dataset to the LLM in the prompt. The LLM then outputs the delta policy parameters.}
    \label{fig:llmirp_method}
\end{figure}

Our goal is ultimately to solve the optimization in Eq.~\ref{eq:policy_min}. We propose to solve this by attempting to iteratively improve our policy parameterization $\theta$. To achieve this, we formulate a \textit{policy improvement operator} $f$, as a mapping that takes in the current best policy parameterization estimate $\theta^i$ and state trajectory $s^i_{1:T}$ and outputs a change to the policy $\Delta \theta^i$:
\begin{equation}
    f(\theta^i, s^i_{1:T}) \to \Delta \theta^i
\end{equation}
The improved policy is then $\theta^{i+1}=\theta^i + \Delta \theta^i$. To fit this operator, we assume access to a dataset $\mathcal{D}=\{d_1, d_2,...,d_N\}$ of policy improvement labels, $d_i=(\theta^i, s^i_{1:T}, \Delta \theta^i)$. We discuss how this dataset can be collected in Sec.~\ref{sec:method-dataset}.

In the typical approach to solving for this data-driven operator, $f$ would be setup as a data-driven model, such as a neural network, which is trained using the dataset $\mathcal{D}$. Here we instead propose to utilize \textit{in-context learning}, leveraging pre-trained LLMs by inputting both our query and the dataset $\mathcal{D}$ into the LLM prompt:
\begin{equation}
    f(\theta^i, s^i_{1:T}, \mathcal{D}) \to \Delta \theta^i
\end{equation}
The result is a sample-efficient method that requires no model design or training, but rather fits the operator all in the forward pass of the pre-trained model.

We utilize the LLM by treating our policy improvement operator as a sequence-to-sequence completion task. The input sequence is the current policy and state trajectory and the output sequence is the policy change. We use the provided dataset to feed example input-output sequences, and the LLM is instructed to complete the sequence for the current policy information to estimate the corresponding policy update:
\begin{equation}
    f(\{\theta^{\mathcal{D}_j}, s^{\mathcal{D}_j}_{1:T},\Delta\theta^{\mathcal{D}_j}\}_{j=1}^k, \theta^i, s_{1:T}^i) \to \Delta \theta^i
\end{equation}

We call our proposed method In-Context Policy Improvement (ICPI) (Fig.~\ref{fig:llmirp_method}). Next, we highlight how we tokenize the policy and state trajectories for input to the LLM and discuss how examples are selected from the dataset to form our query.

\subsection{Tokenization}

To input our policy and state trajectory information, we must tokenize the information to text which the LLM can parse and reason over. 
Following existing in-context learning approaches~\cite{mirchandani_GPM_2023,di_keypoint_2024}, we encode our policy and state information as numeric characters. For our policy and policy update values, $\theta$ and $\Delta \theta$, we simply tokenize each term as characters. 

For the state trajectory $s^i_{1:T}$, we design a task-specific encoding to reflect the error of the execution. In many tasks, the relative change in an outcome is consistent even for differing task parameters. For example, two pucks of different mass may behave differently, but if each is falling short of the goal, we want to push ``harder'' in both cases. Thus, correcting for similar relative errors may involve similar $\Delta \theta$. As such, we encode the state trajectory information as the relative error to the goal, $s^i_{1:T} \to e^i =  s^i_t - \tau_g$. $t$ is selected according to the task. For the $\texttt{slide}$ and $\texttt{roll}$ tasks, $t=T$. For $\texttt{rope-swing}$, $t=\arg \min_t ||s_t - \tau_g||_2$. This relative error is then tokenized as characters.

\subsection{Dataset Example Selection} \label{sec:method-data-selection}

To enable in-context learning, we select a set of example input-output sequences from our dataset $\mathcal{D}$ to be provided along with the current policy execution input. In order to limit the size of the inputs to the LLM and focus the examples on ``useful'' examples, we utilize a K-Nearest Neighbors (KNN) lookup to find similar examples. In particular, we construct a k-D Tree using the vector $v^{\mathcal{D}_j}=[\theta^{\mathcal{D}_j},e^{\mathcal{D}_j}]$, where we normalize the policy and error terms to have comparable scale along each dimension. For a query example $v^i=[\theta^i, e^i]$, we can then perform a KNN lookup to find the most similar examples in our dataset $\mathcal{D}$. Following existing ICL work~\cite{mirchandani_GPM_2023}, we then provide these $k$ examples in order of decreasing distance to the query example. In our experiments, we set $k=20$.

\subsection{Dataset Construction} \label{sec:method-dataset}

Finally, we discuss the creation of our dataset $\mathcal{D}$. For each task instance $\tau \sim P(\tau)$, let $\theta^*$ be a solution to Eq.~\ref{eq:policy_min}. For any nearby execution $\theta$, we can construct the delta label $\Delta \theta = \theta^* - \theta$. $\theta^*$ can be derived several different ways, including by demonstration or via some other algorithmic approach to solving Eq.~\ref{eq:policy_min}, such as Reinforcement Learning or brute force search. Our method then learns to make adjustments based on the results of these laborious methods. This can be seen as a form of algorithm distillation~\cite{laskin2022context}, where we distill the progress of some other algorithm efficiently into our new sequence-to-sequence problem.

In practice, we construct our training data on a per-task basis. For the simulated \texttt{slide} and \texttt{rope-swing} setups, we derive $\theta^*$ using a brute-force search algorithm to find a solution $\theta^*$. For our \texttt{roll-gc-real} task, executing a brute-force search on the real setup would be too slow. Since we are solving a goal conditioned task, we can assign goal labels in hindsight based on the final state of the policy rollout~\cite{pmlr-v100-lynch20a}. We can then generate our dataset by sampling alternative $\theta$ parameterizations, and computing the error to the final state of the ``guiding'' execution.

\section{Results}\label{sec:results}

\subsection{Iterative Policy Improvement}

First, we investigate how our proposed ICPI method performs and provide comparison to baselines. \update{While more sophisticated modeling and learning techniques are state-of-the-art for dynamic manipulation, these methods generally require large-scale datasets, e.g. 54 million examples for rope swing in \citet{chi_irp_2022} or $\sim$5000 steps for Reinforcement Learning of puck sliding in \citet{luo2023sirl}. As such, we focus our comparison to data-efficient baselines and alternate LLM-based methods.}

\subsubsection{Baselines}

\textbf{Random Shooting:} we set up a random shooting method that samples a change to the current policy $\theta^i$ by sampling a randomized offset $\Delta \theta^i \sim \mathcal{N}(0;0.5\cdot c^i \cdot \Delta\theta_{\max})$, where $\Delta\theta_{\max}=\theta_{\max} - \theta_{\min}$.

\textbf{Bayes Opt:} this baseline seeks to directly solve Eq.~\ref{eq:policy_min} as a Bayesian optimization problem~\cite{bayesopt}.

\textbf{KNN-$k$:} this baseline solves the policy iteration by performing only the KNN lookup within $\mathcal{D}$ as described in Sec.~\ref{sec:method-data-selection} and takes the average of the $k$ closest labels. We set $k=5$.

\textbf{Linear KNN-$k$:} this baseline performs the KNN lookup as described in Sec.~\ref{sec:method-data-selection} and fits a linear model to the $k$ closest input-output labels, and inputs $v^i$ to the linear model. This forms a piece-wise linear model of the policy improvement operator. We match ICPI and set $k=20$.

\textbf{In-Context Sequence-Improvement (ICSI):} This method is based on Sequence Improvement technique introduced by \citet{mirchandani_GPM_2023}. In this method, they use cost conditioning to prompt a LLM to improve over previous iterations. We apply their method here by pairing policy parameters $\theta^i$ with their cost $c^i$. We then query the LLM to generate a new policy conditioned on previous $c^i, \theta^i$ examples, and prompting with an improved $c^{i+1}$, asking the LLM to predict the corresponding improved $\theta^{i+1}$.

\textbf{In-Weights Reasoning (IW):} This method seeks to utilize the \textit{in-weights reasoning} capability of LLMs for our tasks. For each task, we describe in natural language the task setup and action space. We also provide the previous executions as $\theta^i, e^i$. We then directly query the model to reason about the system and provide a better policy. This baseline seeks to determine how well \textit{in-weights reasoning} performs on our target tasks.

\subsubsection{Policy Improvement Results}

\begin{table}
    \fontsize{9pt}{9pt}\selectfont
    \centering
    \renewcommand{\arraystretch}{1.5}
    \begin{tabularx}{\linewidth}{c|cc|cc|c}
        \toprule
        Method & \texttt{slide} & \texttt{slide-gc} & \texttt{rope-swing} & \texttt{rope-swing-gc} & \texttt{roll-gc-real} \\
        \hline
        Rand. Shooting & 0.037 (0.050) & 0.077 (0.055) & \textbf{0.007 (0.022)} & 0.004 (0.009) & - \\
        Bayes Opt & 0.054 (0.065) & 0.087 (0.052) & 0.019 (0.018) & 0.014 (0.012) & - \\
        KNN-5 & 0.106 (0.085) & 0.071 (0.043) & 0.020 (0.023) & 0.010 (0.015) & - \\
        Lin. KNN-20 & 0.053 (0.060) & \textbf{0.022 (0.017)} & 0.042 (0.076) & 0.006 (0.017) & 33.824 (28.031) \\
        \hline
        ICSI~\cite{mirchandani_GPM_2023} & 0.030 (0.055) & 0.107 (0.058) & 0.042 (0.067) & 0.021 (0.035) & - \\
        In-Weights & 0.029 (0.055) & 0.102 (0.060) & 0.027 (0.062) & 0.022 (0.037) & - \\
        \hline
        ICPI (Ours) & \textbf{0.013 (0.014)} & 0.025 (0.026) & \textbf{0.007 (0.012)} & \textbf{0.002 (0.006)} & \textbf{17.107 (9.796)} \\
        \bottomrule
    \end{tabularx}
    \caption{Final Best-Policy Mean Performance Comparison $C^{\texttt{task}}_\tau$ ($\downarrow$)}
    \label{tab:quant_final}
\end{table}

\begin{figure}
    \centering
    \begin{subfigure}[t]{0.32\linewidth}
        \centering
        \includegraphics[width=\textwidth]{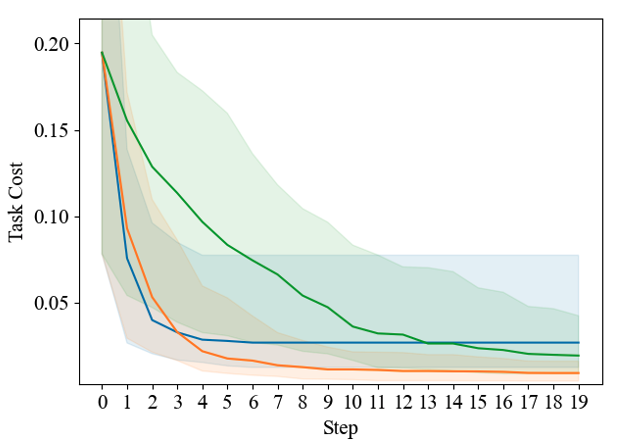}
        \caption{\texttt{slide}}
        \label{fig:slide_task}
    \end{subfigure}
    \begin{subfigure}[t]{0.32\textwidth}
        \centering
        \includegraphics[width=\textwidth]{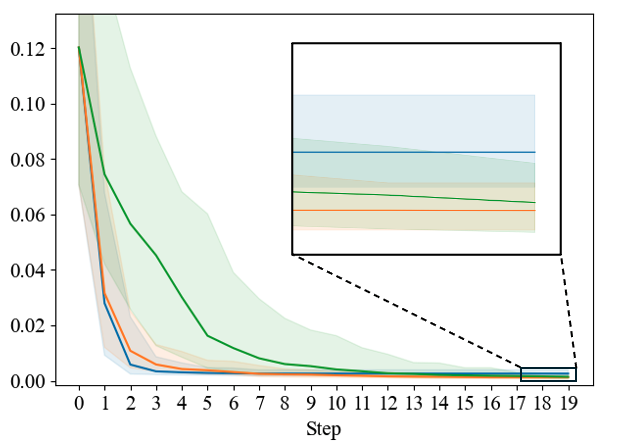}
        \caption{\texttt{rope-swing-gc}}
        \label{fig:rope_swing_gc_task}
    \end{subfigure}
    \begin{subfigure}[t]{0.32\textwidth}
        \centering
        \includegraphics[width=\textwidth]{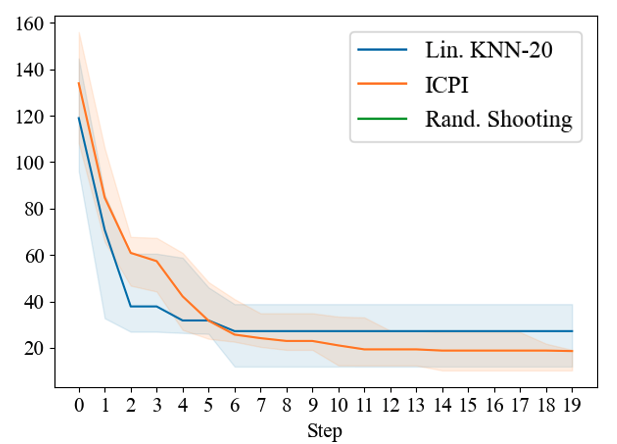}
        \caption{\texttt{roll-gc-real}}
        \label{fig:roll_gc_real_task}
    \end{subfigure}
    \caption{Task Cost convergence plots for the best policy so far at each step across three of our tasks comparing random shooting, piece-wise linear modeling, and our proposed method.}
    \label{fig:task_convergence}
\end{figure}

We generate a dataset $\mathcal{D}$ of $\sim$300 policy improvement examples as described in Sec.~\ref{sec:method-dataset}. For our simulated tasks (\texttt{slide}, \texttt{slide-gc}, \texttt{rope-swing}, \texttt{rope-swing-gc}) we run each policy iteration method on 100 sampled tasks. For our real task (\texttt{roll-gc-real}) we execute 10 sampled tasks and only compare to the piece-wise linear method, as we found it was one of the most competitive methods in simulation. All methods are initialized with the same starting action for fair comparison.

\begin{figure}
    \centering
    \includegraphics[width=0.9\linewidth]{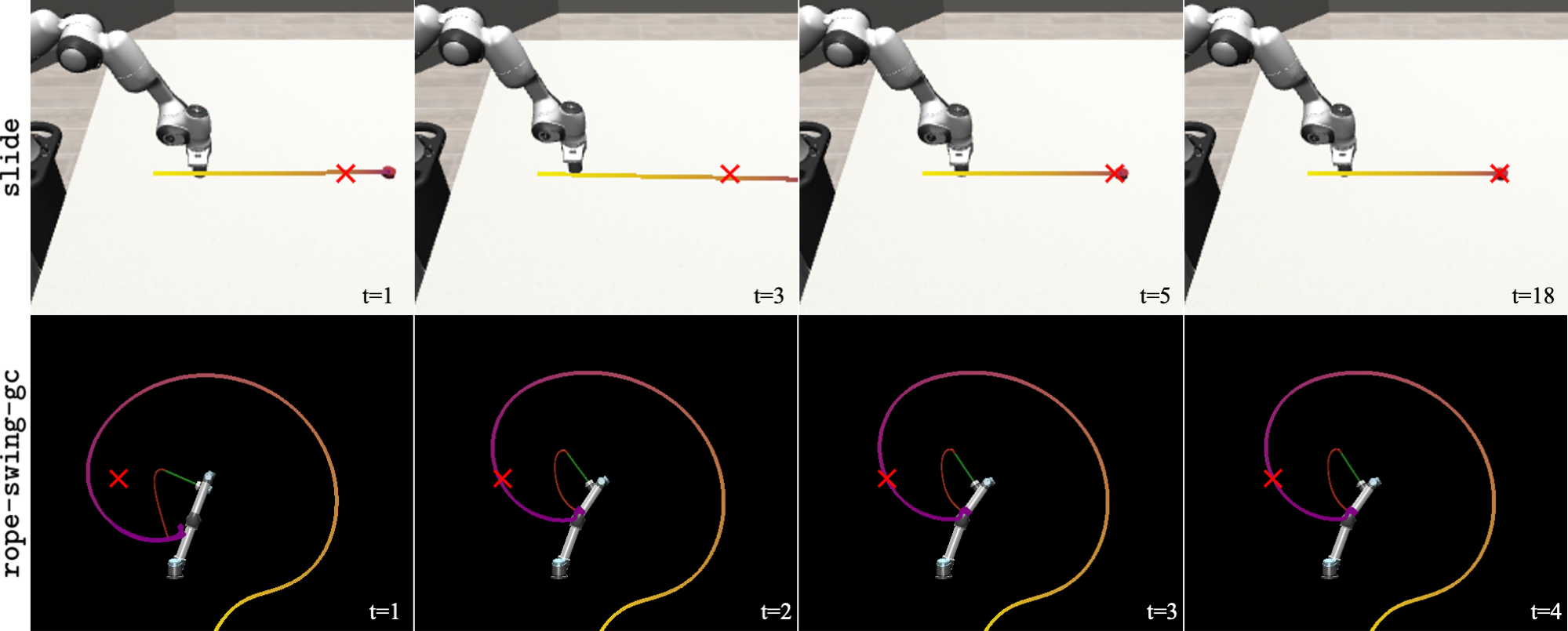}
    \caption{Qualitative examples of iterative in-context policy improvement using our proposed ICPI for the \texttt{slide} and \texttt{rope-swing-gc} tasks.}
    \label{fig:qual_results}
\end{figure}

In Table~\ref{tab:quant_final}, we show the average and standard deviation for the best policy cost after 20 iterations. We found that in most tasks, our proposed ICPI method outperformed the baselines, and was the only method that consistently performed well across all tasks. In Fig.~\ref{fig:task_convergence}, we compare the evolution of the best policy cost as a function of iteration step for random shooting, piece-wise linear (Lin. KNN-20), and our method. We see that ICPI consistently outperforms random samples and converges to a better result than the piece-wise linear approach, including on the real robot. Qualitative policy iteration results for ICPI are shown in Figs.~\ref{fig:teaser} and~\ref{fig:qual_results}.

Our in-context method outperformed utilizing in-weights reasoning on our task. This highlights that while dynamic reasoning may not be readily available in LLMs, their in-context abilities can still be useful. ICPI also outperformed the ICSI approach~\cite{mirchandani_GPM_2023}, showing the value of our proposed policy improvement formulation for unlocking the benefits of in-context learning.

\subsection{Design Ablations} \label{sec:ablations}

We next investigate how design decisions impact ICPI performance. In particular, we look at two variations on our method. In the first variation, we exchange our tokenization method, and instead of directly providing $e^i$, we instead provide both $s^i_t,\tau_g$ to the model. Second, we investigate how the choice in LLM affects our method performance. The comparisons across our simulated tasks are shown in Tab.~\ref{tab:icpi_ablations}. We find that model choice has a notable impact on in-context learning performance. While we are not privy to all model information, we notice that the newer, large-scale reasoning model (\texttt{gpt-4o}) seems to outperform older (\texttt{gpt-3.5-turbo}) and smaller (\texttt{gpt-4o-mini}) models. We also see that directly providing the relative error $e^i$ outperforms providing the state and goal $s_t^i,\tau_g$ separately, even though both contain the same information.

\begin{table}
    \fontsize{9.5pt}{9.5pt}\selectfont
    \centering
    \renewcommand{\arraystretch}{1.5}
    \begin{tabularx}{\linewidth}{cc|cc|cc}
        \toprule
        Inputs & LLM Model & \texttt{slide} & \texttt{slide-gc} & \texttt{rope-swing} & \texttt{rope-swing-gc} \\
        \hline
        $\theta^i, e^i$ & \texttt{gpt-3.5-turbo} & 0.031 (0.045) & 0.033 (0.026) & 0.013 (0.019) & 0.007 (0.014) \\
        $\theta^i, e^i$ & \texttt{gpt-4o-mini} & 0.026 (0.035) & 0.039 (0.031) & 0.010 (0.015) & 0.009 (0.018) \\
        \hline
        $\theta^i, s_t^i,\tau_g$ & \texttt{gpt-4o} & 0.017 (0.022) & 0.055 (0.050) & 0.002 (0.002) & 0.006 (0.012) \\
        \hline
        $\theta^i, e^i$ & \texttt{gpt-4o} & 0.013 (0.014) & 0.025 (0.026) & 0.007 (0.012) & 0.002 (0.006) \\
        \bottomrule
    \end{tabularx}
    \caption{ICPI Ablations on Final Best-Policy Mean Performance Comparison $C^{\texttt{task}}_\tau$ ($\downarrow$). Last row is settings used for our experiments.}
    \label{tab:icpi_ablations}
\end{table}

\section{Discussion}\label{sec:discussion}

We proposed In-Context Policy Improvement (ICPI), utilizing pre-trained Large Language Models (LLMs) for sample-efficient dynamic manipulation policy improvement. We demonstrate that while LLMs struggle to apply \textit{in-weights} reasoning to dynamic tasks, \textit{in-context learning} outperforms alternative policy iteration methods, both in simulation and on a real robot. This work adds to a growing body of work showing the utility of in-context learning in robotics~\cite{mirchandani_GPM_2023,di_keypoint_2024}, which we hope motivates further investigation, \update{both utilizing non-robotics and robotics specific transformer models.}


\subsection{Limitations}

A limitation of our proposed method is the overhead of utilizing large pretrained models. These models incur both computational, financial, and environmental overhead~\cite{bender_too_big} and the low-level details of models are not available, which can make it difficult to understand model choice impact. Our method relies on a policy improvement dataset, which may be difficult to collect for complex tasks. 
Self-play~\cite{pmlr-v100-lynch20a} and demonstrations~\cite{di_keypoint_2024} could be utilized to ease the data collection burden.

In this work, found that some amount of feature selection was important to task success (see drop in performance when providing $s^i,\tau_g$ vs. $e^i$ in Sec.~\ref{sec:ablations}). This seems to indicate that in-context learning may not yet be capable of the feature learning ubiquitous in modern machine learning. While signs indicate that in-context learning may continue to improve with improved LLMs (see \texttt{gpt-4o} vs. \texttt{gpt-3.5-turbo} in Sec.~\ref{sec:ablations} and \citet{mirchandani_GPM_2023}), it is unclear if more advanced pattern-based reasoning will be achievable purely in-context. Additionally, we only investigated relatively low-dimensional input-output formats; while low-dimensional parametric actions are common in dynamic manipulation, they can limit the dexterity of the system. \update{Our results present sufficient features for our proposed tasks - we leave to future work an extensive study of task dimensionality and feature extraction for in-context learning.}


\bibliography{main}

\newpage
\appendix

\section{Appendix}

\subsection{Example Prompts}

We provide example prompts and responses for our method along with the other LLM baselines. The examples provided are for the simulated \texttt{slide} task.

\subsubsection{In-Context Policy Improvement (Ours)}

See Sec.~\ref{sec:method} for in-depth discussion of how we construct our prompts. We note the task-agnostic header, which is fixed across our experiments.

\fbox{
\parbox{\textwidth}{
\ttfamily\small
\textcolor{purple}{Prompt:}\\
You are a pattern generator machine. I will give you a series of patterns with INPUTS and OUTPUTS as examples. Then you will receive a new INPUTS, and you have to generate OUTPUTS following the pattern that appears in the data.
\\\\
Patterns are provided per-line as: INPUTS;OUTPUTS
\\\\
Only reply with an estimate for the OUTPUTS. OUTPUTS should be 3 values separated by spaces.
\\\\
-0.061 0.242 0.771 -0.001 -0.017;-0.022 0.002 -0.065

0.050 0.284 0.830 0.027 0.001;0.024 -0.007 -0.059

\ldots

0.065 0.276 0.723 -0.086 0.002;0.009 0.001 0.049

-0.031 0.252 0.672 -0.076 -0.011;0.039 -0.036 -0.137
\\\\
-0.016 0.269 0.780 -0.155 -0.004;
\\\\
\textcolor{orange}{Response:}\\
0.004 -0.011 -0.060
}
}

\subsubsection{In-Context Sequence-Improvement~\cite{mirchandani_GPM_2023}}

Following \citet{mirchandani_GPM_2023}, we perform sequence improvement in-context. We follow their proposed cost prompting strategy of prompting with a randomly sampled improved cost.

\fbox{
\parbox{\textwidth}{
\ttfamily\small
\textcolor{purple}{Prompt:}\\
You are a pattern generator machine. I will give you a series of patterns with INPUTS and OUTPUTS as examples. Then you will receive a new INPUTS, and you have to generate OUTPUTS following the pattern that appears in the data.
\\\\
Patterns are provided per-line as: INPUTS;OUTPUTS
\\\\
Only reply with an estimate for the OUTPUTS. OUTPUTS should be 3 values separated by spaces.
\\\\
0.550;0.153 0.112 0.812

0.528;0.464 0.112 0.500

\ldots

0.094;0.298 0.350 0.825

0.044;0.153 0.269 0.825
\\\\
0.000;
\\\\
\textcolor{orange}{Response:}\\
0.153 0.112 0.825
}
}

\subsubsection{In-Weights Reasoning}

Our final LLM baseline seeks to employ the \textit{in-weights} capabilities of the model. As such, we provide a detailed description of each task, setting, and action space, along with previous interactions. The descriptions for the other tasks are provided in similar specificity to attempt to ground the in-weights reasoning to the task.

\fbox{
\parbox{\textwidth}{
\ttfamily\small
\textcolor{purple}{Prompt:}\\
You are a planner responsible for providing an action for a robot to execute. You should reason carefully about the physics of the scenario when planning. You may receive a description of previous interactions with the target system and should consider these previous interactions when deciding how to act. Each task is solved via a single action execution - it does not take multiple actions to complete the task.

Only reply with the action the robot should execute, as a space-separated list of numbers.

A robot is positioned in front of a table. The robot has an arm which is extended over the table. The tool connected to the robot is a cylinder, starting perpendicular to the tabletop. In front of the robot arm on the table is a puck, which is also a cylinder, and it can move freely on the tabletop. The task is to use the robot arm to slide the puck on the table to a goal position. The robot tool and the puck start so that planar motions over the tabletop will cause the two to collide (i.e., no vertical motion of the robot tool is required). The robot has a single action to complete the task - it cannot take multiple actions.

The robot tool starts at location (0.000, 0.000) on the tabletop. The puck starts at location (0.100, 0.000) on the tabletop. The goal is to move the puck to (0.800, 0.000).

The action space of the robot is the polar coordinates for a planar motion of the robot tool relative to its starting position and the time it takes to complete the motion. The first term is the angle of motion, where 0 degrees moves straight forward along the x dimension. The second term is the distance in that direction the robot will move. The third term is the time it will take to complete the motion.

The actions are bounded. The angle of motion must lie between -0.464 and 0.464 radians. The distance must be between 0.112 and 0.350 meters. The motion can take between 0.500 and 5.000 seconds to complete. The provided actions will be clipped into this range. The action is mapped to a dense set of tool locations via linear interpolation and executed on the system.

The robot has already interacted with the system several times. You will receive a set of descriptions of these interactions. You should consider the previous interactions and how to improve upon them when selecting the next action to attempt.

The previous interactions will be provided with one interaction per line. Each line will describe the interaction as the action taken and the final position of the puck relative to the goal location, separated by a semi-colon. Each item is provided as a space-separated list of values.
\\\\
0.000 0.350 0.780;-0.654 -0.000

0.000 0.350 0.780;-0.654 -0.000

\ldots

0.000 0.269 0.780;-0.162 -0.000

-0.016 0.269 0.780;-0.155 -0.004
\\\\
\textcolor{orange}{Response:}\\
0.000 0.350 0.780
}
}

\end{document}